# Interpolation, Extrapolation, Hyperpolation: Generalising into new dimensions

Toby Ord[1]

This paper introduces the concept of *hyperpolation*: a way of generalising from a limited set of data points that is a peer to the more familiar concepts of interpolation and extrapolation. Hyperpolation is the task of estimating the value of a function at new locations that lie outside the subspace (or manifold) of the existing data. We shall see that hyperpolation is possible and explore its links to creativity in the arts and sciences. We will also examine the role of hyperpolation in machine learning and suggest that the lack of fundamental creativity in current AI systems is deeply connected to their limited ability to hyperpolate.

## Introduction

Interpolation and extrapolation are two kinds of generalisation: ways of applying an idea in a broader domain than we've seen so far. Interpolation asks what lies between the examples we've already seen, while extrapolation asks what lies beyond.

They are widely used concepts that find technical application within science and engineering (especially in data science, numerical analysis, machine learning, economics, and computer graphics), where many different mathematical methods for interpolation and extrapolation are used. These concepts are also used in a less formal manner in many other areas, such as philosophy, art, and futurism, where we might make more qualitative interpolations and extrapolations. For instance, we might ask whether an emerging kind of music is mainly an interpolation between two existing genres; or whether affordable land travel, sea travel, and air travel are likely to be followed by affordable space travel.

In all these contexts, interpolation and extrapolation are seen as twin concepts; an inseparable pair. I want to suggest that what appear to be twins are in fact two triplets — that they have a hitherto unknown sibling. Once you meet this new member of the family, you will see how the three concepts fit together cleanly into a coherent whole, with each governing its own realm of generalisation. And you will see how recognising this long-lost triplet might improve our understanding of creativity — in art, in science, in evolution, and in AI.

---

[1] I'd like to thank Daron Acemoglu, Anders Sandberg, Fin Moorhouse, Phil Trammell, Loren Fryxell, Matthew van der Merwe, Owen Cotton-Barratt, Sebastian Farquhar, Thomas Moynihan, and Thomas Homer-Dixon for insightful discussion and comments.

Let us begin by considering a simple case, where all our known examples lie along a common 1-dimensional line. We could think of this as knowing the value of $f(x)$ at some finite set of points, $\{x_i\}$, and wanting to generalise to understand the value of $f(x)$ at new points. If those new points lie between two existing data points (and so are within the shortest line segment spanning the $\{x_i\}$), then this is an instance of interpolation. If they lie along that same line but outside this line segment, it is extrapolation.

But what if they don't lie on the line at all?

Suppose that the underlying space of possibilities is 2- or 3-dimensional but that the data points all lie on a common line that cuts through that space. Cases like this where the existing examples are restricted to a lower-dimensional subspace within the wider space of possibilities are very common (especially in machine learning). In such situations, attempting to generalise away from the line containing all our data points would be neither interpolation nor extrapolation. We shall call it *hyperpolation*.[2] Where interpolation looks between the known examples and extrapolation looks past them, hyperpolation *transcends* them — treating them as a special case within a higher dimensional structure.[3]

It is not immediately obvious that hyperpolation is even possible — that one can say *anything* useful about what lies in a direction orthogonal to all the known examples. Yet we shall see that it *is* possible, exploring a variety of examples.

---

[2] I am using the prefix 'hyper-' in its sense of 'beyond' — in much the same way as 'hyperspace' (a space with more than the usual 3 dimensions) and 'hyperplane' (a generalisation of a 2D plane to higher dimensions).

[3] Since writing this essay, I've learned of two closely related approaches:
   Margaret Boden (2007) distinguishes three kinds of creativity: combinational, exploratory, and transformational. While not spelled out formally, the last of these appears to be closely connected with hyperpolation (especially generative hyperpolation). For example, it is 'done by altering, adding, or deleting dimensions of the original conceptual space'.
   As part of an ingenious psychology experiment to show how humans generalise, Lucas et al (2012) introduce the concept of 'superspace extrapolation', which I take to be identical to hyperpolation. Their paper demonstrates that people *can* hyperpolate and do so in a way that is compatible with Bayesian updating on a hypothesis space of mathematically simple functional forms.

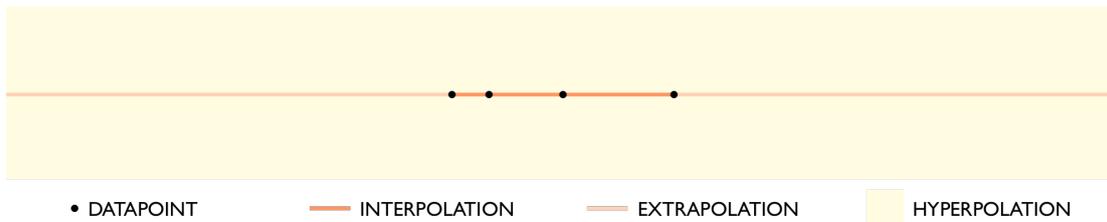

*Figure 1.* The different domains in which one performs interpolation, extrapolation, and hyperpolation — for the case where the data lies on a common 1D line within a larger 2D space.

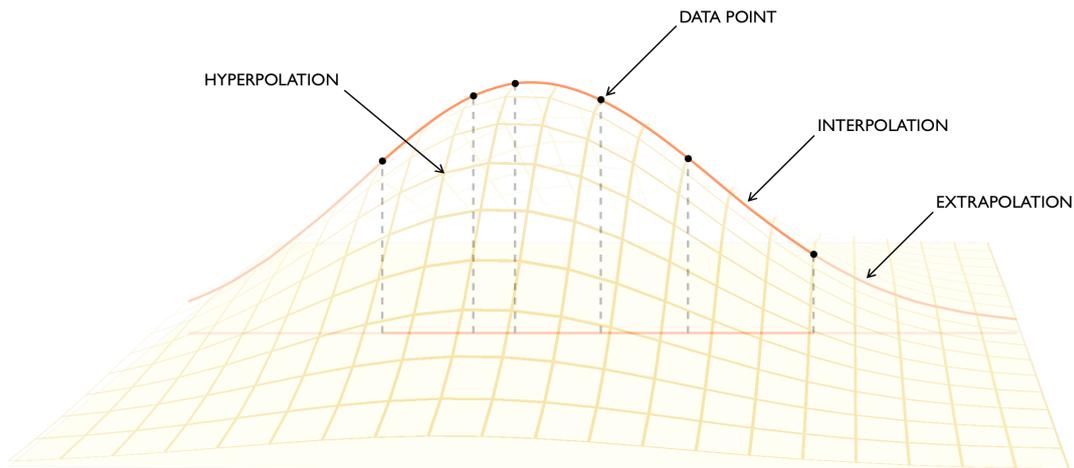

*Figure 2.* We can now look at the value of $f(x)$ at each data point, representing it as a height above the plane from *Figure 1*. We can interpolate curves between the data points and extrapolate curves going off towards $-\infty$ and $+\infty$, providing estimates of $f(x)$ everywhere along the subspace of the data. Finally we can hyperpolate a surface (shown as the golden mesh) that extends this curve into another dimension, estimating values of $f(x)$ everywhere in the full space.

It is important to note that while we will see that hyperpolation is possible, it is not a mathematically 'well-posed problem'. There are infinitely many ways of generalising the function off the slice where the data lies with no canonical way of narrowing this down to a single best generalisation.

However this is also true for interpolation and extrapolation, and it doesn't prevent those being meaningful concepts that we use formally and informally in everyday life. Just as we tame the infinity of possible solutions to interpolation and extrapolation through constraints such as the smoothness of the function, a preference for simplicity, or outputting a probability distribution over answers — so we can for hyperpolation. Nonetheless it is worth bearing in mind that as with interpolation and extrapolation, we will be searching for 'reasonable answers', 'answers that work well in practice', or 'best answers given a prior for the shape of the function' rather than canonical solutions.

## Defining hyperpolation

Let's start with a formal definition of function generalisation and then see how it divides neatly into three realms.

We consider an unknown function $f$ from $\mathbb{R}^n$ to $\mathbb{R}$. We start with a set of data points of the form $(x_i, f(x_i))$ which provide examples of the value of $f$ at different locations, $\{x_i\}$, in its domain.[4] Let's say that the problem of *function generalisation* is to use this information to predict the value of $f(x)$ for new points. So function generalisation is a problem where you take $\{(x_i, f(x_i))\} \in \mathcal{P}(\mathbb{R}^n \times \mathbb{R})$ and some $x \in \mathbb{R}^n$ which isn't a member of $\{x_i\}$ as inputs and you output a single real number. An algorithm succeeds at this problem to the extent that its outputs are close to the ground truth value of $f(x)$.[5]

To understand the different kinds of function generalisation, it is helpful to use the following standard definitions:

A *convex combination* of the points $\{x_i\}$ is any weighted sum of them, $x = \Sigma_i \alpha_i x_i$, where each $\alpha_i \in [0, 1]$ and $\Sigma_i \alpha_i = 1$. A convex combination thus is a kind of mixture of the original points. The *convex hull* of the points is the set of all their convex combinations. In 1-dimension, a convex hull will be a line segment; in 2-dimensions a convex polygon; in 3-dimensions a convex polyhedron; and in general a convex polytope.

An *affine combination* of the points $\{x_i\}$ is any weighted sum of them $x = \Sigma_i \alpha_i x_i$, where each $\alpha_i \in \mathbb{R}$ and $\Sigma_i \alpha_i = 1$. An affine combination could lie between the original points, or beyond them, but can only head out in directions determined by the points themselves. The *affine hull* of the points is the set of all their affine combinations. A 1-dimensional affine hull forms a line; a 2-dimensional affine hull forms a plane; a 3-dimensional affine hull forms a 3D space. It will always be an infinite flat space, which is known as an *affine space* or when it is a part of a larger space, an *affine subspace*. It is closely related to the span of a set of vectors (a vector subspace) but doesn't have to include the origin.

We can now carve the problem of function generalisation into three parts:

1) *Interpolation* is when $x$ is in the convex hull of $\{x_i\}$.
2) *Extrapolation* is when $x$ is not in the convex hull of $\{x_i\}$, but is in the affine hull.
3) *Hyperpolation* is when $x$ is in neither the convex nor affine hull of $\{x_i\}$.

---

[4] The bold $x$ denotes a point in the *n*-dimensional space, while the regular $x$, $y$, $z$ etc. denote the locations within individual dimensions.

[5] Note that there are many ways one could measure such success. For example, it will depend on what set of functions you try to generalise and on how you penalise different patterns of deviation from the ground truth.

These three cases are mutually exclusive and jointly exhaustive: every instance of function generalisation is exactly one of interpolation, extrapolation, or hyperpolation.

One remaining question is whether we take the data points to be sacrosanct. There are really two different versions of function generalisation which take different approaches to this question. The strict version treats the data points as ground truth, and so requires the generalised function $f(x)$ to exactly pass through every data point. The flexible version merely requires $f(x)$ to go close to the data points. It penalises solutions that stray further from the data but allows this closeness of fit to be traded off against a superior functional form for $f(x)$, such as one that is simple, smooth, or more likely *a priori*. This is often used in contexts where the data includes some form of statistical noise.

Sometimes the terms 'interpolation' and 'extrapolation' are restricted to the strict version, while the flexible version is referred to as 'curve fitting' or 'regression'. But much of what we will cover here applies to both versions so we will use the terms 'interpolation' and 'extrapolation' inclusively.

With these two versions in mind, we can note that if we were to relax the definition of function generalisation so that it applied to predicting $f(x)$ for *any $x$* — even a value of $x$ that is already in the data set — then there would be a fourth category of function generalisation:

   0) *Autopolation* is when $x$ is in $\{x_i\}$.

On the strict version, this is a trivial kind of generalisation, recapitulating a data point we have already seen. But on the flexible version, one could make use of the other data points to better understand the underlying function and thereby improve upon the noisy estimate given by the existing data point at $x$. Indeed, if we are allowing for noise, we might even want to allow multiple data points at the same location in the domain, $x_i$, but with different values of $f(x_i)$.

So far, we've defined hyperpolation as generalising from a set of data points in a subspace of a domain to the parts of the domain beyond that subspace. We could also consider related cases where we start with a function of $n$ variables and a data set whose affine hull covers its entire domain, then generalise this to a function of $n$+1 variables. For example, we could start with a function of one variable, $f(x)$, and a dozen data points in its domain and then generalise to a function $f(x, y)$ and try to estimate its value at various combinations of $x$ and $y$.

This is a useful broadening of the idea of hyperpolation, though it introduces some additional issues. For example, we can only perform this kind of hyperpolation on real world data if the previously omitted variable was constant in all our data points — otherwise the apparent variation of $f$ in response to $x$ may have been entirely driven by the omitted $y$.

## Can we hyperpolate?

Consider the curve below (*Figure 3*). Suppose scientists observed such a curve with great precision — with a set of data points so finely spaced as to leave us in little doubt that this smooth interpolation is the true shape. What could they say about how this function might behave if its domain were extended into a second dimension: from a line into a plane? In other words, if we think of this curve as a vertical cross section of a 2D surface, what can we say about the shape of that surface?

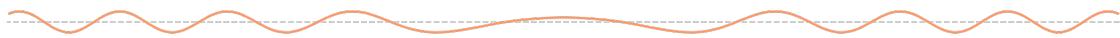

*Figure 3*. Can we hyperpolate this curve into a surface?

It is not immediately obvious how to go about extending this curve into a surface. One simple option that is always present is to just extrude it into the new dimension. If we call this curve *f(x)*, we could just set *f(x, y) = f(x)*. If we truly knew the value of *f(x)* everywhere along the original line, then this would be a generalisation of nearest-neighbour interpolation/extrapolation, that we could call *nearest-neighbour hyperpolation*.[6] Another simple approach would be to assume the curve behaves the same way in the new coordinate: *f(x, y) = f(x) + f(y)*. Or we could choose something more complex — with an uncountable infinity of possibilities.

Rather than taking a generic approach that could apply regardless of the functional form of the curve, let's look at this particular curve more closely to see if it is special in a way that might reveal the best approach to take. The curve itself is a smooth sinuous shape. It is somewhat like a sine wave. More precisely, it is like a cosine — for it is symmetrical around the *y*-axis. Indeed, supposing we have access to more data points to the left or right, we would observe that the curve approaches cos(*x*) as *x* approaches both –∞ and +∞. If a group of scientists were determined to find its underlying equation, then through trial and error they could eventually arrive at the rather simple equation:

$$f(x) = \cos(\sqrt{x^2 + 400})$$

How might this be generalisable to a function of two variables? Interestingly, we can do so without adding complexity by taking it as an opportunity to remove the constant term 400. For example, we could replace '400' with '*y*' and take this curve as being the slice through:

$$f(x, y) = \cos(\sqrt{x^2 + y})$$

along the line where *y* = 400. Or even better, we could replace the 400 with *y*², noting that then we have the ubiquitous mathematical motif $\sqrt{x^2 + y^2}$, representing a Euclidean distance. Now we have the candidate equation:

$$f(x, y) = \cos(\sqrt{x^2 + y^2})$$

---

[6] If we don't know the value of *f(x)* everywhere on the line, then this splits into two different methods: to use the value of the nearest known point, even when that has a different *x*-coordinate, or to first interpolate and extrapolate to get estimated values everywhere on the line, then to use those for the points off the line at the same *x*-coordinate.

This can be written even more compactly as:

$$f(r) = \cos(r)$$

We can plot this surface, cutting it along the line *y* = –20 to reveal our original curve as a cross section:

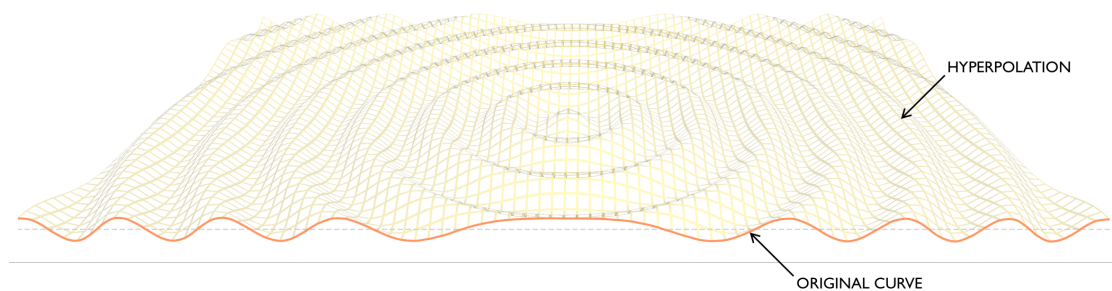

*Figure 4*. The original curve (orange) and its hyperpolation as *f(r)* = cos(*r*) (yellow).

This shows that our original curve can be expressed very simply as a cross section of a ripple pattern, with circular symmetry. Indeed, we have probably all seen something very like the original curve as a momentary shape on the wall of a pool after a small object has been dropped in. And I presume some people could jump straight to this hyperpolation immediately after being shown the original curve.[7]

While a definitive answer would require a full theory of mathematical simplicity, I conjecture that this is the uniquely best hyperpolation of the original curve. Or almost unique. There is an extra wrinkle due to symmetry that can arise depending on the details of how we formulate the problem. For there is also a symmetrical solution reflected around our original line, and we may not be able to break the symmetry as to which one to use. If we can't break the symmetry, we might think of the hyperpolation as providing a distribution of possible surfaces, with these two being the equal highest likelihood answers. Alternatively, if we were already working in a 2D domain and knew that all the original data points were of the form (*x*, –20), then we'd have strong reason to think that this was the unique correct solution, as the solution where the centre of the pattern is at (0, 0) is simpler than the one where the centre is at (0, –40), whose equation is:

$$f(x, y) = \cos(\sqrt{x^2 + (y + 40)^2})$$

---

[7] The best answer I've heard yet (out of two colleagues I've posed the challenge to) is: 'Don't tell me, I think I can see what it is. It's something rotationally symmetric with a pattern of ridges. Something like sin($x^2 + y^2$) or cos. It's cos. Oh, that's strange, the central peak is lower… — oh, this might just be off-axis. I think it is cos($x^2 + y^2$).'

Let's try another example. Can we hyperpolate a hyperbola? Consider the hyperbola:

$$f(x) = \sqrt{x^2 + 1}$$

How could we most naturally extend it out into a 2D surface? Or, equivalently, what 2D surface is a hyperbola most likely to be a cross-section of?

An extremely simple hypothesis is that the hyperbola is a vertical slice of a cone:

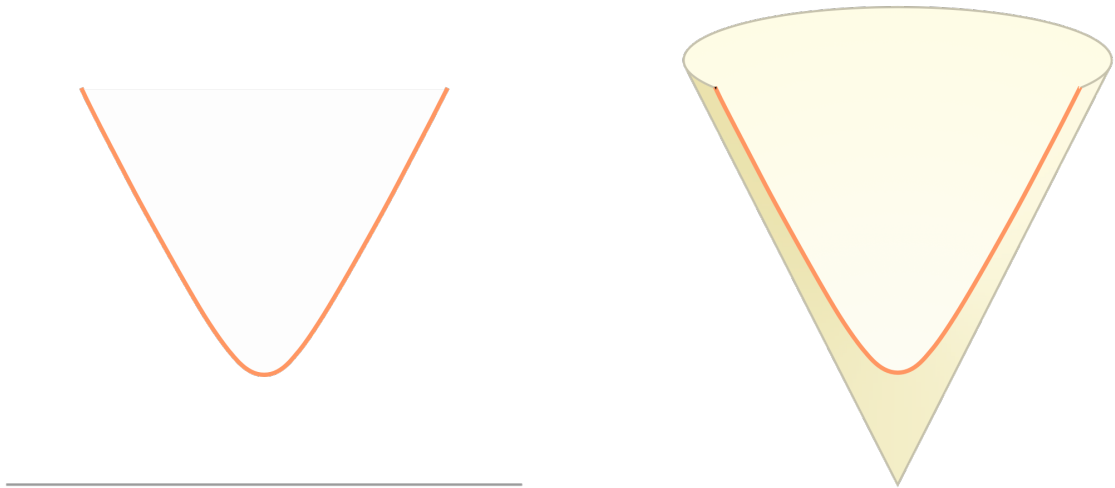

*Figure 5*. The hyperbola as a vertical slice of the cone, *f(r)* = *r*.

Perhaps you have even performed this generalisation yourself — noticing the hyperbolic patch of light that a lamp casts upon a wall and hyperpolating out the entire cone of light that must be invisibly filling the space, allowing you to easily predict the shape of light cast on any surfaces brought near or to predict whether a given point in space would be illuminated were you to hold an object there.

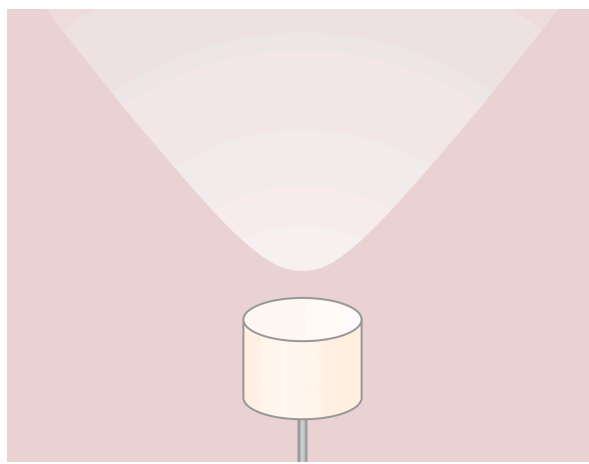

*Figure 6*. A hyperbola as a vertical slice of a cone: *f(r)* = *r*.

Mathematically, we could view this case either as hyperpolating the 1D boundary between the light and dark parts of the wall into a 2D boundary (the surface of a cone), or as hyperpolating the 2D patch of light into the 3D volume of light that makes up the interior of the cone.

In both of these examples, the full hyperpolated shape (the ripple or the cone) is in some important sense *simpler* than the slice. It is a more primitive mathematical object, with fewer arbitrary parameters. Even if you were just shown the data on the slice and were not told of the larger space it lay within nor asked to hyperpolate, you perhaps should come up with a theory to explain the data that involves the simpler higher-dimensional shape.

One place we explicitly see this kind of reasoning is within theoretical physics. For example cosmologists noticed that if there were a continuum of universes in which the physical constants took on each possible value (a multiverse), then there would no longer any need to explain why our constants took on a rare combination of values which could allow for observers like us. We'd only be able to arise on the slices compatible with intelligent life. Similarly, quantum theorists noticed that some of the weirdness of the Copenhagen interpretation could be avoided if we assumed that our universe was just one part of a multiverse containing 'worlds' where all possible quantum outcomes are realised.[8]

There is also a connection to the concept of *abduction* (or *inference to the best explanation*) from the philosophy of science.[9] This is a kind of inference where you examine the evidence at hand and search for a simple hypothesis which would explain that evidence. Even though this hypothesis doesn't follow from the evidence by mere logic alone, if the hypothesised event were likely enough *a priori* and would also be likely to produce the evidence, then the evidence can provide reason to believe it happened. For example, we may find shards of glass on a tiled floor and hypothesize that these were all from a single drinking glass which fell and shattered. We cannot logically deduce that, but it is likely enough *a priori* and would be likely to produce the evidence we see.

---

[8] For an approachable explanation of the development and purpose of such multiverse theories, see Tegmark (2014).

[9] See for example, Boyd (1981).

In our case the evidence would be a set of data points that lie in a common subspace. These can lead us to hypothesize a simple higher-dimensional surface which would explain those data points. For example to hypothesize a cone of light to explain a hyperbolic patch of light. One might therefore view hyperpolation as being justified via abduction, or alternatively, to use hyperpolation as a mathematical model for explaining abduction.

This connection with abduction also suggests a close connection with Bayesian inference. Indeed interpolation, extrapolation, and hyperpolation can all be viewed as Bayesian inference. In each case you start with some prior distribution for $f(x)$ then update it based the evidence contained in the data points (the values $f$ takes at each of the $\{x_i\}$).[10] This gives a posterior distribution over functions which have been selected for both *a priori* likelihood and their fit with the evidence. You then apply all of these to $x$, to get a distribution of values that $f(x)$ could take. This could be compressed down to a single value by taking the value with the highest likelihood, or by taking a weighted average. Alternatively, one could simply output the distribution of values itself. Viewed at this level of abstraction through the lens of Bayesian inference, the method is exactly the same whether we are performing interpolation, extrapolation, or hyperpolation.

A key feature of the ripple and cone examples is that there exists a higher dimensional function which is simpler than the starting slice. This seems important to the theory of hyperpolation. If $f(x, y)$ were more complex than the slice of it lying within the subspace then one wouldn't have much hope of pinning it down (there would be too many different shapes of $f(x, y)$ at that level of complexity which are compatible with the evidence in the slice). Something similar is true for interpolation and extrapolation too. There are infinitely many functions that pass through the data points, but fewer simpler ones, so you only have much hope of specifying the function precisely if it is amongst the simplest functions that fit the data. It is possible that the cases where the whole is simpler than the slice are the only cases where there is a maximum likelihood answer that isn't just nearest-neighbour hyperpolation.

Both examples involve hyperpolated surfaces with a circular symmetry. This greatly helps with finding a unique hyperpolation. Hyperpolation is easier when the subspace containing the data cuts through the higher dimensional structure of *f(x)*, and much harder when it doesn't. For example in the ripple example, *f(x)* has a curved structure and the straight line containing the data cuts through this structure at many different radii as well as many different angles from the origin. So while this line only sees one value of *y*, it sees many values of *r* and *θ*. This allows the data to provide evidence that the angle is irrelevant (implying a circularly symmetric structure) as well as evidence about what that structure is.

---

[10] If the data points are taken to be perfectly accurate, we get the strict version of hyperpolation, if they are taken to include noise, we get the flexible version.

A similar process could also work when there was independent structure in an *x* direction and a *y* direction so long as the subspace of the data ran diagonally to these, cutting through both kinds of structure (e.g. if *f(x, y)* = *xy*, and the {$x_i$} run along a diagonal line).

Finally, let's consider an example with a less clean solution. You are presented with photograph and asked to imagine what would happen next. In other words, to treat it as a frame in a film and to imagine what would be on the subsequent frames. This is a complex case of hyperpolation, where we have a 2D pixel grid (which we could model as a function from locations in $\mathbb{R}^2$ to RGB colours in $[0, 1]^3$) and we are being asked to hyperpolate it in the time dimension.

If we were given multiple frames of a film, we could extrapolate these over time, but here we just have what amounts to a single frame, where all data points have the same value in the time dimension. So we can only estimate pixel values at future times via hyperpolation.

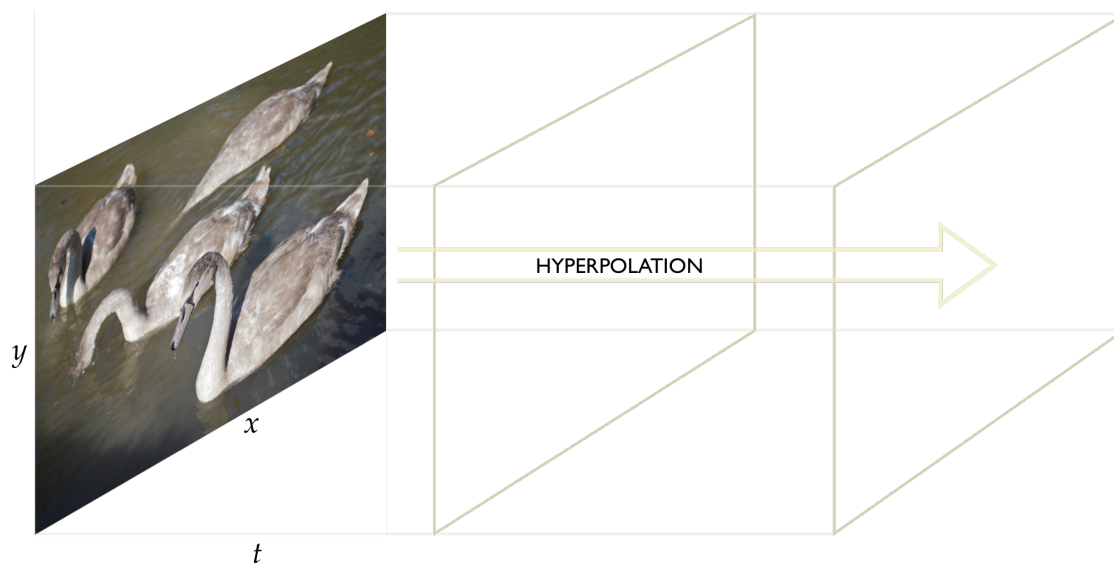

*Figure 7.* The hyperpolation of an image into the time dimension would give a video showing how that scene may develop over time. It is not extrapolation because the new locations at which we are being asked to estimate pixel values are outside the subspace of the existing data.

There is clearly no single correct answer here — no unique simplest continuation given by a universal theory of mathematical simplicity. Instead, one is being asked to say something about the probability distribution of the future sequences of frames, or to give the maximum likelihood sequence, or to sample from the distribution, or something like that.

Interestingly, this is a case where we can break the symmetry that sometimes prevents a unique hyperpolation. There is something objectively better about a hyperpolation whereas time increases, autumn leaves detach from the tree and fall to the ground, rather than one in which they spontaneously rise from the ground and attach themselves to the tree. We are using our prior knowledge that macroscopic systems have an arrow of time (the direction of increasing entropy) in order to break the symmetry in this hyperpolated dimension.

Indeed, much of what we are doing here relies on detailed priors about the temporal behaviour of pixel grids (or about the behaviour of the 3D objects that give rise to these patterns of pixels). These priors are partly a product of our evolution and partly of our lifetime's experience. While they don't allow us to predict the future of the scene with high reliability, we do astronomically better than chance (almost all sequences of future frames just immediately degenerate into random static).

Given these examples where we can and do hyperpolate, it is interesting to return to the original intuition many people have that hyperpolation would be impossible. One thought that may be driving this is that hyperpolation looks superficially like extrapolating into the new dimension, but since there is at most one data point on any line that extends into the new dimension, this would be extrapolation from a single data point — a recipe for failure.

However, as we've seen, hyperpolation can use many different data points on the subspace of the data to help predict a point off that subspace. For example, in the ripple case, if we want to estimate $f(x_1, y)$ (at a familiar $x$-coordinate, but a novel $y$-coordinate) we use many data points with differing $x$-coordinates to understand the overall functional form, then use that to estimate the value of $f(x_1, y)$. This way that many different data points from across the data subspace participate in estimating a value that lies off that subspace is one of the most interesting features of hyperpolation.

## Machine Learning & Hyperpolation

Function generalisation is a key part of machine learning. The paradigm of *supervised learning* involves training a model on a set of data points of the form $\{(x_i, f(x_i))\}$. The model learns some form of internal approximate representation of $f(x)$ which can then be queried to estimate the value of $f(x)$ at an arbitrary new point, $x$.

A common use-case is *classification*, in which *f(x)* takes on a discrete set of values: for example, True and False, or a finite set of different animals the model is being trained to detect. But supervised learning can also be used to learn functions that take on a continuous set of values such as $\mathbb{R}$ or $\mathbb{R}^n$. And even when *f(x)* takes on only discrete values, the model is often viewed as learning a continuous function: the shape of the decision boundary between the regions where *f(x)* takes on its different values, or the degree to which the input fits each of the discrete classification classes.

There are a wide variety of machine learning techniques for supervised learning. These range from simple methods such as nearest-neighbour, to more complex techniques such as support vector machines or neural networks.

Machine learning models perform at their best when the new point being queried, *x*, is 'in distribution' — that is, it appears to have been sampled from the domain by the same process that chose the training data. This means the models tend to do better in cases of interpolation (because points that are outside the convex hull of all existing data are more likely to be out of distribution) and extrapolation (because points that lie off the subspace containing the training data are even more likely to be out of distribution). But distance from the training data also matters. As with all methods of interpolation and extrapolation, it is generally more accurate when the new point is near an existing data point in a densely sampled region.

The distribution of the training data is often thought of in terms of a *latent space*. This is a lower dimensional manifold that sits within the full space of potential inputs and contains most (or all) of the training data points. For example, if one plotted the set of all GPS coordinates of mobile phones in 3D space, you would see they fall within a curved 2D manifold within that space: the surface of the Earth. The *manifold hypothesis* is the claim that even when the space of potential inputs is very high-dimensional, natural data tends to fall on a much lower-dimensional manifold. This is a common working assumption with both theoretical and empirical support.[11]

It often observed that models tend to fail badly when applied to points that don't lie on the manifold containing the training data — a particular kind of out-of-distribution failure. And we could understand this as the models having difficulty hyperpolating out of this manifold.

This would require a slight generalisation of my definition of hyperpolation. I've defined it within a flat space, but one could also define it within a (possibly curved) manifold. If so, the analogues of interpolation and extrapolation would need to curve with the manifold (e.g. by being defined in terms of geodesics within the manifold) and the analogue of hyperpolation would need to go in a direction orthogonal to the data manifold. For example, if the manifold were the surface of a sphere, interpolation and extrapolation[12] would go along great circles within that surface and hyperpolation would go to spheres of different radii.

---

[11] For example, see Cayton (2005) for a survey and Fefferman et al. (2016) for empirical investigation.

It would be interesting to use the concept of hyperpolation to explore the abilities of different kinds of machine learning models to generalise beyond the subspace (or manifold) of the data. For example:

- How good are different machine learning architectures at hyperpolating?[13]
- How much worse are they at hyperpolation compared with interpolation or extrapolation?
- How can we make them better at hyperpolation?
- Could any existing architectures hyperpolate well enough to reconstruct the cone or ripple pattern?
- How does the accuracy of hyperpolation drop off with distance from the data?
- Is it better to interpolate and extrapolate first and then try to hyperpolate from that?
- Is it practically important that the slice through the data has *zero* thickness in the new dimension, or would a few new examples just off the existing subspace not help much?

**Generative models & creativity**

What about generative AI?

While there have been dramatic improvements in generative AI, the systems in 2024 still seem to have trouble transcending their training data. The best models can now produce text, images, or songs that we would call creative if produced by one of our friends or family members. But it is not clear this is the right standard, as these models are also trained on an inhumanly vast amount of data, and assessments of creativity depend on the context of what else the creator has seen. A painting or story that we first see as highly creative may seem much less so if we find out that it falls within a genre or style we hadn't yet encountered and that its creator had access to hundreds of similar examples while making it.

---

[12] The distinction between interpolation and extrapolation may break down on some manifolds. For instance, any polygon drawn on a sphere divides it into two finite parts and it is not obvious which one is the inside (for interpolation) vs the outside (for extrapolation). The difference between interpolation and extrapolation here would appear to be more one of degree than one of kind.

[13] For example, Gaussian processes are defined to explicitly have a prior over $f(x)$ which they update on the training data, and they are able to represent long-range structure such as periodic functions. Would that make them better equipped than basic neural networks to hyperpolate the ripple pattern?

Even if we could determine that the generated output wasn't similar to any one of the training examples, we might still feel that it was a kind of interpolation between them. For instance, that it involved elements from multiple existing works, but didn't really bring anything new to the table. Or that it provided a good parody of Oscar Wilde by exaggerating his style, but was really just an extrapolation of things we had already seen.

Of course many great works by humans also involve remixing and extrapolating what we've seen already. But we also possess the ability to transcend everything that has come before — to take a step off the beaten path in a new direction, orthogonal to the subspace of existing works. Think of Marcel Duchamp's *Fountain*, Allen Ginsberg's *Howl*, or The Beatles' *Sgt. Pepper's Lonely Hearts Club Band*. And this kind of originality is something that hasn't yet been clearly demonstrated by even the best generative models. While it is hard to measure this to know for sure, one suggestive thing is the apparent lack of *any* examples of foundation models producing new results in maths or science.

This suggests that the language of interpolation, extrapolation, and hyperpolation could also be useful for thinking about generative AI and creativity. To do so, we need to be able to express what a generative model does in terms of function generalisation. A natural way to do this is to view a generative model as possessing an implicit quality function, $q(x)$, that assesses the quality of an item (such as an image, sound, or piece of text). If so, then the task of generative AI is not just to find *any* new images, sounds, or text, that lie off the data manifold, but to find new items which of high quality. This is possible without explicit training on $q(x)$ if one assumes that the training data provides a set of examples with unusually high quality (i.e. much higher than that of a random point in the input space).

One gloss we could put on this distinction is that hyperpolation in science is about making accurate predictions away from the subspace where all our data lies, while hyperpolation in art is about finding new high-quality locations away from that subspace of all previous examples.

But the truth is somewhat messier than this. Hyperpolation in science isn't just about being able to predict something new — the ability to understand the functional form that lies behind the predictions is important in and of itself. For example, when we see the entirety of the symmetrical ripple pattern, we gain a deeper explanation of the sinuous curve we had observed. This could be of value even if we never use it to make predictions beyond the original subspace. Multiverse theories are a nice example: their value doesn't lie in making predictions about what is happening in other universes, but in explaining why the physical constants take certain values in our universe.

Moreover, progress in the sciences also benefits from the second style of hyperpolation. Ideas like Mendelian inheritance, quantum mechanics, and non-Euclidean geometry are not mere interpolations or extrapolations of existing ideas. They expanded the space of scientific ideas in fundamentally new directions. And much of the scientific work that we consider highly creative is also about finding new high-quality ideas that lie outside the existing subspace of ideas.

In science, in art, and in machine learning, it is easiest to assess the quality of new locations (and thus find the high-quality regions enabling the generation of good new content) when the locations are: close to a known example, interpolations of known examples, or extrapolations of known examples. It is much harder to assess the quality of new locations when they are off the subspace (or manifold) of the training data. Doing so would be the generative version of hyperpolation, and difficulty achieving this is reflected in the paucity of generated content that really transcends the training data.

On this view, a big thing that is limiting the kinds of creativity that foundation models display (and thus the kinds of creative tasks they can achieve) is a lack of ability at hyperpolation.

But it is worth noting that there *are* generative AI systems that excel at certain kinds of hyperpolation tasks.[14] Recall the task of looking at the pixels that make up a single photograph and predicting pixel values at future times. There are generative models that do this, taking an image as input and outputting a movie.[15] They have been trained on sequences of movie frames, with the goal being to predict later frames based only on the initial frame.[16]

---

[14] While these are generative AI systems, they may be solving the standard hyperpolation problem, rather than the generative type we just considered. In particular, it doesn't seem like we need to invoke the idea of a quality function to explain what is going on. While they generate novel videos, it may be more like trying to predict the objective ground truth future that follows from a given frame than it is like trying to produce a high-quality image or song.

[15] For example, Blattmann et al. (2023) and Xu et al. (2024).

[16] Relatedly, AI systems can also be trained to fill in missing internal patches of obscured images (interpolation) or expand cropped images into full images (extrapolation).

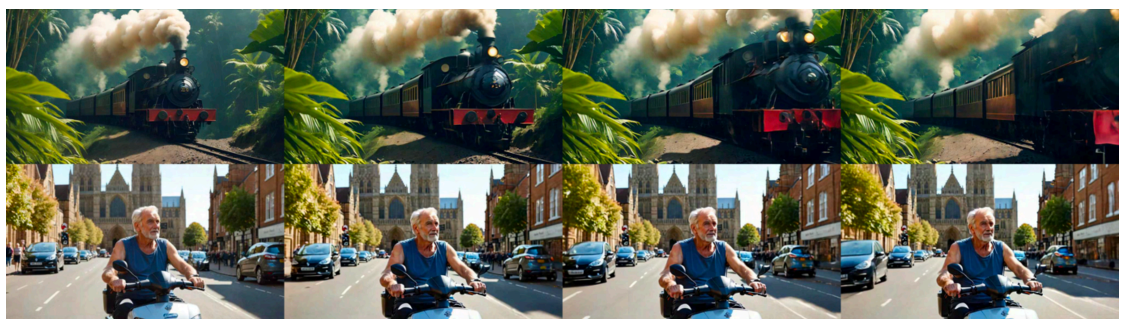

*Figure 8.* Two examples of image hyperpolation from StabilityAI's Stable Video Diffusion (figure reproduced from Blattmann et al. (2023)). The input images on the far left have been hyperpolated forward in time by a few seconds to create a plausible video, of which three new frames are shown. Note how the train correctly moves forward in time with the smoke remaining still. In the second example, the cars on the left are moving forward towards the camera, while those on the right are moving forward away from it. The camera remains a fixed distance from the rider, but is clearly moving backwards compared to the rest of the scene, leading to more of the scene coming into view.

Such systems start with simple implicit priors about the grid of pixel colours over time and in the learning process update these to much more complex priors that must effectively include things like rotations of 3D objects and their projections down to 2D configurations of pixels, as well as many particular kinds of objects such as people, animals, or cars, and their typical physical behaviours.

How do these systems hyperpolate so competently when others have such trouble? The answer lies in the structure of the learning problem. Most machine learning systems learn *a generalisation*. These image-to-video systems learn *to generalise*. So they are not hyperpolating the training data (a set of images with their associated videos) but are hyperpolating the input (a 2D grid of pixels with their associated colours). It is not that each training data-point was a pixel location at a time, with its corresponding colour, and the model extending this to new times — it is that each training data-point was an entire coloured pixel grid along with its corresponding hyperpolation. It is not that it is learning some arbitrary function $f(x)$ and we want it to hyperpolate that function, but that the particular $f(x)$ it's learning is *the hyperpolation function for still images*.[17]

I'm not sure what the lesson is here for improving the abilities of machine learning systems to hyperpolate out of the data manifold. Perhaps it is that we would do better training specialised systems to learn how to hyperpolate particular kinds of things from supervised examples, than in training general foundation models and hoping that they transcend their training data.

---

[17] Which it is presumably learning mostly through interpolation and extrapolation of the training set.

Now that we've seen how we can understand hyperpolation for generative AI in terms of generalising a quality function, we might also ask: What about other generative systems, beyond AI?

**Evolution & hyperpolation**

When we look around the natural world, we see a remarkable range of seemingly creative solutions to the challenges in a species' life. We see advanced organs like eyes, wings, and brains; as well as complex behaviours like tool-making or social insects. Yet three billion years ago, there was none of that. All life was single-celled. The bewildering complexity of life in our post–Cambrian explosion world — where trillions of cells join together in specialised roles to create macroscopic organisms — was still far in the future. While it is hard to state precisely, it seems safe to say that these complex organs and behaviours of today's natural world are off the subspace of life three billion years ago. Yet here they are. How was evolution able to get to somewhere truly new?

We can split this into two questions that I think have different answers:

1) How was evolution able to even conceive of, or specify, these new phenotypes that are so different to everything that had come before?
2) How was evolution able to assess their quality, to find not just any radically new phenotypes, but those that solved the challenges the species were facing?

The answer to the first question comes from the connection between the genotype and the phenotype. The genotype can be thought of as an organism's entire sequence of base pairs of DNA, each of which can be a C, G, A, or T. This forms a vast discrete space of genotypes with a number of dimensions equal to the number of base pairs (from about 1 million dimensions for a simple bacterium, to 1 billion dimensions for a human, all the way to 100 billion dimensions for the longest known genotype). Mutation is a process that can change one or more these base-pairs, leading to a new genotype that is different on one or more dimension. Mutation can also insert or delete base pairs, changing the length of the DNA sequence.[18] There are thus paths consisting of successive mutations that could (in theory) take evolution anywhere in this space of genotypes.

In contrast, the phenotype of an organism is its combination of observable traits and characteristics. Where the genotype is a simple (but immense) abstract space of sequences of symbols, the phenotype is a much more concrete set of physical characteristics (including the chemical, biological, and psychological characteristics that are set by that physical arrangement). While changes to an organism's genotype are discrete changes in one or more dimensions, the corresponding changes to an organism's phenotype are more complex and continuous. For example, scientists have discovered about 50 different genes and genetic regions that contribute to determining a human's adult height.[19]

---

[18] There are other, more complicated kinds of mutation, such as duplication of parts of the genome that don't need to concern us here.

We can thus think of this as something like a billion-dimensional abstract discrete space of genotypes that (in combination with other environmental and epigenetic factors) maps to a somewhat lower dimensional and more continuous space of phenotypes. The answer to how evolution can even 'conceive of' or specify a new phenotype that is off the subspace of existing phenotypes is via their genotypes. The abstract, digital, genotypes give a way of specifying points in the much messier, concrete, continuous space of phenotypes. Where one might look at a set of species and have no idea how to even think of something that is off that subspace of phenotypes, it is much easier to specify one by changing random base pairs scattered throughout the genotype.

Of course if we do make random changes to a genotype, they will be unlikely to lead to a 'creative' solution to one of a species' problems. We still face the second challenge of how we find out the quality of a new phenotype. The answer is of course that the new phenotype is tested in the crucible of the environment. If it leads to better survival and reproduction (better evolutionary fitness) then it spreads and forms the basis for future mutations. If it doesn't, it tends to die out.

So evolution finds successful new phenotypes that are far off the subspace of existing organisms by a sequence of small steps, each of which involves a change to the genotype, which is turned into a new phenotype, which is tested by the environment. While there are many other complications in a full story of how evolution hyperpolates, the mechanism of genetic mutation is sufficient to dispel the initial mystery of how evolution could possibly find high quality locations that lie in a direction orthogonal to every existing phenotype. Indeed, it makes it feel natural and inevitable.

It is, however, worth saying a few words about another major evolutionary mechanism: sexual reproduction. Sexual reproduction produces offspring based on a combination of the genotypes of the parents. It is reminiscent of interpolation between the two parents, but unlike true interpolation, there isn't a 1-dimensional continuum of possible children. Instead, traits from each parent can be brought together in many combinations, giving each pair of parents a multidimensional space of possible children. I'm not sure if sexual reproduction could hyperpolate without assistance from mutation, but we do know that the reverse is true. At my hypothetical starting time of three billion years ago, there was no sexual reproduction. That entire reproduction mechanism was outside the subspace of existing phenotypes, yet mutation was able to discover it.

---

[19] McEvoy & Visscher (2009).

**Evolution & human creativity**

It is possible that there are many parallels here to the way humans generate creative new ideas or styles which aren't a mere interpolation or extrapolation of what has come before. In many cases we may only be able to take a single step in the completely new direction, relying on repetition of this process to reach any distant locations. And the human who takes that first step may not even be able to reliably judge the quality of the new location — they may need to test it, seeing whether a new invention is profitable, a new style of song is popular, or a new idea is accepted. If so, that new thing may get widely adopted and become the starting point for further steps. And once there is a step or two in a new direction, that might provide enough of a run-up to extrapolate in that new direction, leaping ahead to much more distant locations.

Sometimes this process of iteration happens very quickly with tiny steps found by semi-random exploration: such as trying variations of a theme at the piano, with further development of those that seemed promising; or *automatism*, where the artist paints with unconscious movements in the hope of interesting chance developments.

And it is suggestive that many areas of human creativity have a digital expression: an abstract combinatorial language of letters, words, or musical symbols that map to a concrete continuous world of meanings or sounds. In such cases, one can use evolution's trick of easily specifying novel locations in this abstract space, even when you can't easily specify novel locations in the concrete space. William S. Burroughs popularised such a technique for making poetry from random words cut from a newspaper, allowing the poet to quickly reach strange ideas and styles that were alien to anything that had come before.[20] Musicians have used this technique for their lyrics in avant-garde albums such as David Bowie's *Diamond Dogs* and Radiohead's *Kid A* — semi-randomly manipulating the syntactic representation of language until they find novel semantic ideas that catch their interest.[21]

---

[20] One might at first think of merely sampling and reordering words from one source would just be an interpolation of its ideas, but I don't think this is so. For instance, one could explain Darwin's theory of evolution using just words from the King James Bible, but I think it is fair to say the idea of evolution transcends the set of ideas found in the Bible.

[21] In the 1975 BBC documentary *Cracked Actor*, Bowie explained: 'What I've used it for more than anything else is igniting anything that might be in my imagination — only it can often come out with very interesting attitudes to look into.'

One could even see other 'constraints that breed creativity' in the same way. Take the traditional poetic constraints of meter and rhyme. These act on a syntactic level by restricting word choice based on the sounds of the words rather than their meanings. So we find rhymed words that are related in sound, but often with a surprising juxtaposition of meanings.

**Conclusions**

The purpose of this essay was to introduce the idea of hyperpolation and begin to explore its possibilities. We saw that hyperpolation can be mathematically defined as a counterpart to interpolation and extrapolation, with each one governing generalisation in a different part of the domain. And we saw that hyperpolation is possible, both in abstract mathematical cases like the ripple pattern and practical cases like projecting time forwards from a still image. We explored its connections to machine learning — especially to the challenges in generalising off the data-manifold in supervised learning and to generating items that are outside the subspace of existing ideas or genres in generative AI. Finally, we explored how evolution overcomes these challenges to generate new phenotypes that break free of the subspace of all those that have come before. Throughout, we noted the rich connections between hyperpolation and creativity, whether artistic or scientific.

But all of this has barely scratched the surface of what appears to be a deep and important concept. There are many natural pathways for further exploration:

- Developing a menu of practical methods for hyperpolation.[22]
- Analysing the abilities of different machine learning paradigms to hyperpolate beyond their training data.
- Developing generative models with different modalities that learn to hyperpolate.
- Characterising the set of circumstances where hyperpolation can be fruitfully applied.[23]
- Exploring hyperpolation of a data set to multiple new dimensions simultaneously.
- Exploring the philosophical connections to abduction.

---

[22] e.g. Nearest-neighbour hyperpolation and other analogues to interpolation and extrapolation methods; replacement of constants with variables; introducing new variables.

[23] e.g. Does the functional form of the underlying function need to be simpler (or as simple) as the functional form restricted to the data subspace? How does the presence of noise affect this?

For me, the most exciting possibility is the connection to creativity. Can a better understanding of the technical concept of hyperpolation shed light on those flashes of insight where we realise that everything we were looking at is just one aspect of a greater whole? Can it help with the quest to find fundamentally new ideas, genres, mechanisms, or styles that transcend everything that came before?